\pdfoutput=1

\documentclass[11pt]{article}

\usepackage{acl}

\usepackage{times}
\usepackage{latexsym}

\usepackage[T1]{fontenc}

\usepackage[utf8]{inputenc}

\usepackage{microtype}
\usepackage{amsmath}
\usepackage{amsfonts}
\usepackage{tikz}
\usetikzlibrary{positioning, calc, decorations.pathreplacing}
\usepackage{booktabs}
\usepackage{caption}
\usepackage{subcaption}
\usepackage{cleveref}

\definecolor{casblue}{RGB}{8,62,140}
\definecolor{casred}{RGB}{203,65,84} 
\definecolor{casgreen}{RGB}{34,139,34} 
\definecolor{casyellow}{RGB}{218,165,32} 
\definecolor{caspurple}{RGB}{148,0,211} 
\mathchardef \mhyphen = "2D

\newcommand{\softmax}{\operatorname{softmax}}

\title{Boundary Smoothing for Named Entity Recognition}

\author{Enwei Zhu\textsuperscript{\rm 1,2} \and Jinpeng Li\textsuperscript{\rm 1,2,\thanks{~~Corresponding author.}}  \\
        \textsuperscript{\rm 1}HwaMei Hospital, University of Chinese Academy of Sciences \\
        \textsuperscript{\rm 2}Ningbo Institute of Life and Health Industry, University of Chinese Academy of Sciences \\ 
    	\texttt{\{zhuenwei,lijinpeng\}@ucas.ac.cn}}

\begin{document}
\maketitle
\begin{abstract}
Neural named entity recognition (NER) models may easily encounter the over-confidence issue, which degrades the performance and calibration. Inspired by label smoothing and driven by the ambiguity of boundary annotation in NER engineering, we propose \emph{boundary smoothing} as a regularization technique for span-based neural NER models. It re-assigns entity probabilities from annotated spans to the surrounding ones. Built on a simple but strong baseline, our model achieves results better than or competitive with previous state-of-the-art systems on eight well-known NER benchmarks.\footnote{Our code is available at \url{https://github.com/syuoni/eznlp}.} Further empirical analysis suggests that boundary smoothing effectively mitigates over-confidence, improves model calibration, and brings flatter neural minima and more smoothed loss landscapes. 
\end{abstract}

\section{Introduction} \label{sec:intro}
Named entity recognition (NER) is one of the fundamental natural language processing (NLP) tasks with extensive investigations. As a common setting, an entity is regarded as correctly recognized only if its type and two boundaries exactly match the ground truth. 

The annotation of boundaries is more ambiguous, error-prone, and raises more inconsistencies than entity types. For example, the CoNLL 2003 task contains four entity types (i.e., person, location, organization, miscellaneous), which are easy to distinguish between. However, the boundaries of a entity mention could be ambiguous, because of the ``boundary words'' (e.g., articles or modifiers). Considerable efforts are required to specify the ``gold standard practice'' case by case. Table~\ref{tab:conll2003-examples} presents some examples from CoNLL 2003 Annotation Guidelines.\footnote{\url{https://www-nlpir.nist.gov/related_projects/muc/proceedings/ne_task.html}.} In addition, some studies have also reported that incorrect boundary is a major source of entity recognition error \citep{wang-etal-2019-crossweigh,eberts2019span}. 

\begin{table}[t]
    \centering \small
    \begin{tabular}{ll}
        \toprule
        Text & Boundary words \\
        \midrule
        \textcolor{red}{[}The \textcolor{blue}{[}White House\textcolor{blue}{]$_\texttt{ORG}$}\textcolor{red}{]$_\texttt{ORG}$} & Article \\
        \midrule
        \textcolor{blue}{[}The \textcolor{red}{[}Godfather\textcolor{red}{]$_\texttt{PER}$}\textcolor{blue}{]$_\texttt{PER}$} & Article \\
        \midrule
        \textcolor{red}{[}\textcolor{blue}{[}Clinton\textcolor{blue}{]$_\texttt{PER}$} government\textcolor{red}{]$_\texttt{ORG}$} & Modifier \\
        \midrule
        \textcolor{red}{[}Mr. \textcolor{blue}{[}Harry Schearer\textcolor{blue}{]$_\texttt{PER}$}\textcolor{red}{]$_\texttt{PER}$} & Person title \\
        \midrule
        \textcolor{blue}{[}\textcolor{red}{[}John Doe\textcolor{red}{]$_\texttt{PER}$}, Jr.\textcolor{blue}{]$_\texttt{PER}$} & Name appositive \\
        \bottomrule
    \end{tabular}
    \caption{Examples of CoNLL 2003 Annotation Guidelines and potential alternatives. The gold annotations are marked in \textcolor{blue}{blue [*]}, whereas the alternative annotations are in \textcolor{red}{red [*]}.}
    \label{tab:conll2003-examples}
\end{table}

Recently, span-based models have gained much popularity in NER studies, and achieved state-of-the-art (SOTA) results \citep{eberts2019span,yu-etal-2020-named,li-etal-2021-span}. This approach typically enumerates all candidate spans and classifies them into entity types (including a ``non-entity'' type); the annotated spans are scarce and assigned with full probability to be an entity, whereas all other spans are assigned with zero probability. This creates noticeable \emph{sharpness} between the classification targets of adjacent spans, and may thus plague the trainability of neural networks. In addition, empirical evidence shows that these models easily encounter the \emph{over-confidence} issue, i.e., the confidence of a predicted entity is much higher than its correctness probability. This is a manifestation of miscalibration \citep{guo2017calibration}. 

Inspired by label smoothing \citep{szegedy2016rethinking,muller2019when}, we propose boundary smoothing as a regularization technique for span-based neural NER models. By explicitly re-allocating entity probabilities from annotated spans to the surrounding ones, boundary smoothing can effectively mitigate over-confidence, and result in consistently better performance. 

Specifically, our baseline employs the contextualized embeddings from a pretrained Transformer of a \texttt{base} size (768 hidden size, 12 layers), and the biaffine decoder proposed by \citet{yu-etal-2020-named}. With boundary smoothing, our model outperforms previous SOTA on four English NER datasets (CoNLL 2003, OntoNotes 5, ACE 2004 and ACE 2005) and two Chinese datasets (Weibo NER and Resume NER), and achieves competitive results on other two Chinese datasets (OntoNotes 4 and MSRA). Such extensive experiments support the effectiveness and robustness of our proposed technique. 

In addition, we show that boundary smoothing can help the trained NER models to preserve calibration, such that the produced confidences can better represent the precision rate of a predicted entity. This corresponds to the effect of label smoothing on the image classification task \citep{muller2019when}. Further, visualization results qualitatively suggest that boundary smoothing can lead to flatter solutions and more smoothed loss landscapes, which are typically associated with better generalization and trainability \citep{hochreiter1997flat,li2018visualizing}.

\section{Related Work} \label{sec:related-work}
\paragraph{Named Entity Recognition} The mainstream NER systems are designed to recognize flat entities and based on a sequence tagging framework. \citet{collobert2011natural} introduced the linear-chain conditional random field (CRF) into neural network-based sequence tagging models, which can explicitly encode the transition likelihoods between adjacent tags. Many researchers followed this work, and employed LSTM as the encoder. In addition, character-level representations are typically used for English tasks \citep{huang2015bidirectional, lample-etal-2016-neural, ma-hovy-2016-end, chiu-nichols-2016-named}, whereas lexicon information is helpful for Chinese NER \citep{zhang-yang-2018-chinese, ma-etal-2020-simplify, li-etal-2020-flat}. 

Nested NER allows a token to belong to multiple entities, which conflicts with the plain sequence tagging framework. \citet{ju-etal-2018-neural} proposed to use stacked LSTM-CRFs to predict from inner to outer entities. \citet{strakova-etal-2019-neural} concatenated the BILOU tags for each token inside the nested entities, which allows the LSTM-CRF to work as for flat entities. \citet{li-etal-2020-unified} reformulated nested NER as a machine reading comprehension task. \citet{shen-etal-2021-locate} proposed to recognize nested entities by the two-stage object detection method widely used in computer vision. 

Recent years, a body of literature emerged on span-based models, which were compatible with both flat and nested entities, and achieved SOTA performance \citep{eberts2019span,yu-etal-2020-named,li-etal-2021-span}. These models typically enumerate all possible candidate text spans and then classify each span into entity types. In this work, the biaffine model \citep{yu-etal-2020-named} is chosen and re-implemented with slight modifications as our baseline, because of its high performance and compatibility with boundary smoothing. 

In addition, pretrained language models, also known as contextualized embeddings, were also widely introduced to NER models, and significantly boosted the model performance \citep{peters-etal-2018-deep, devlin-etal-2019-bert}. They are used in our baseline by default. 

\paragraph{Label Smoothing} \citet{szegedy2016rethinking} proposed the label smoothing as a regularization technique to improve the accuracy of the Inception networks on ImageNet. By explicitly assigning a small probability to non-ground-truth labels, label smoothing can prevent the models from becoming too confident about the predictions, and thus improve generalization. It turned out to be a useful alternative to the standard cross entropy loss, and has been widely adopted to fight against the over-confidence \citep{zoph2018learning,chorowski2017towards,vaswani2017attention}, improve the model calibration \citep{muller2019when}, and de-noise incorrect labels \citep{lukasik2020does}. 

Our proposed boundary smoothing applies the smoothing technique to entity boundaries, rather than labels. This is driven by the observation that entity boundaries are more ambiguous and inconsistent to annotate in NER engineering.\footnote{We note that \citet{shen-etal-2021-locate} also allocate a weight to the non-entity but partially matched spans; however, boundary smoothing additionally regularizes the weight of entity spans, which is intuitively crucial for mitigating over-confidence.} To the best of our knowledge, this study is the first that focuses on the effect of smoothing regularization on NER models.

\section{Methods} \label{sec:methods}
\subsection{Biaffine Decoder}
A neural network-based NER model typically encodes the input tokens to a sequence of representations $x = x_1, x_2, \dots, x_T$ of length $T$, and then decodes these representations to task outputs, i.e., a list of entities specified by types and boundaries. 

We follow \citet{yu-etal-2020-named} and use the biaffine decoder. Specifically, the representations $x$ are separately affined by two feedforward networks, resulting in two representations $h^s \in \mathbb{R}^{T \times d}$ and $h^e \in \mathbb{R}^{T \times d}$, which correspond to the start and end positions of spans. For $c$ entity types (a ``non-entity'' type included), given a span starting at the $i$-th token and ending at the $j$-th token, a scoring vector $r_{ij} \in \mathbb{R}^c$ can be computed as: 
\begin{equation}
    r_{ij} = (h^s_i)^{\mathrm T} U h^e_j + W (h^s_i \oplus h^e_j \oplus w_{j-i}) + b,
\end{equation}
where $w_{j-i} \in \mathbb{R}^{d_w}$ is the $(j-i)$-th width embedding from a dedicated learnable matrix; $U \in \mathbb{R}^{d \times c \times d}$, $W \in \mathbb{R}^{c \times (2d+d_w)}$ and $b \in \mathbb{R}^c$ are learnable parameters. $r_{ij}$ is then fed into a softmax layer: 
\begin{equation}
    \hat{y}_{ij} = \softmax (r_{ij}), 
\end{equation}
which yields the predicted probabilities over all entity types. 

The ground truth $y_{ij} \in \mathbb{R}^c$ is an one-hot encoded vector, with value being 1 if the index corresponds with the annotated entity type, and 0 otherwise. Thus, the model can be optimized by the standard cross entropy loss for all candidate spans: 
\begin{equation}
    \mathcal{L}_\mathrm{CE} = -\sum_{0 \leq i \leq j < T} y_{ij}^{\mathrm T} \log(\hat{y}_{ij}).
\end{equation}

In the inference time, the spans predicted to be ``non-entity'' are first discarded, and the remaining ones are ranked by their predictive confidences. Spans with lower confidences would also be discarded if they clash with the boundaries of spans with higher confidences. Refer to \citet{yu-etal-2020-named} for more details.

\begin{figure}[t]
    \centering
    \begin{subfigure}{\columnwidth}
    \centering
    \begin{tikzpicture}[scale=0.5, transform shape]
    \foreach \i in {0,1,...,10}{
        \draw[dotted] (0, -\i) -- (10, -\i);
        \draw[dotted] (\i, 0) -- (\i, -10);
    }
    \node[left, font=\LARGE] at (-0.5, -5) {Start};
    \node[below, font=\LARGE] at (5, -10.75) {End};
    \foreach \i in {0,1,...,9}{
         \node[left, font=\LARGE] at (0, -\i-0.5) {\i};
         \node[below, font=\LARGE] at (\i+0.5, -10) {\i};
    }
    \foreach \i in {0,1,...,9}{
        \foreach \j in {\i,...,9}{
            \draw[rounded corners] (\j, -\i) rectangle (\j+1, -\i-1);
        }
    }
    \draw[rounded corners, fill=casred!100] (2, -1) rectangle (3, -2);
    \draw[rounded corners, fill=casblue!100] (7, -3) rectangle (8, -4); 
    \end{tikzpicture}
    \caption{Hard boundary}
    \label{subfig:hard-boundaries}
    \end{subfigure}
    \begin{subfigure}{\columnwidth}
    \centering
    \begin{tikzpicture}[scale=0.5, transform shape]
    \foreach \i in {0,1,...,10}{
        \draw[dotted] (0, -\i) -- (10, -\i);
        \draw[dotted] (\i, 0) -- (\i, -10);
    }
    \node[left, font=\LARGE] at (-0.5, -5) {Start};
    \node[below, font=\LARGE] at (5, -10.75) {End};
    \foreach \i in {0,1,...,9}{
         \node[left, font=\LARGE] at (0, -\i-0.5) {\i};
         \node[below, font=\LARGE] at (\i+0.5, -10) {\i};
    }
    \foreach \i in {0,1,...,9}{
        \foreach \j in {\i,...,9}{
            \draw[rounded corners] (\j, -\i) rectangle (\j+1, -\i-1);
        }
    }
    \draw[rounded corners, fill=casred!80] (2, -1) rectangle (3, -2);
    \foreach \xs / \ys in {-1/0, 0/1, 1/0, 0/-1}{
        \draw[rounded corners, fill=casred!50] (2+\xs, -1+\ys)  rectangle (3+\xs, -2+\ys);
    }
    \draw[rounded corners, fill=casblue!80] (7, -3) rectangle (8, -4); 
    \foreach \xs / \ys in {-1/0, 0/1, 1/0, 0/-1}{
        \draw[rounded corners, fill=casblue!50] (7+\xs, -3+\ys)  rectangle (8+\xs, -4+\ys);
    }
    \foreach \xs / \ys in {-2/0, -1/1, 0/2, 1/1, 2/0, 1/-1, 0/-2, -1/-1}{
        \draw[rounded corners, fill=casblue!20] (7+\xs, -3+\ys)  rectangle (8+\xs, -4+\ys);
    }
    \end{tikzpicture}
    \caption{Smoothed boundary}
    \label{subfig:smoothed-boundaries}
    \end{subfigure}
    \caption{An example of hard and smoothed boundaries. The example sentence has ten tokens and two entities of spans (1, 2) and (3, 7), colored in red and blue, respectively. The first subfigure presents the entity recognition targets of hard boundaries. The second subfigure presents the corresponding targets of smoothed boundaries, where the span (1, 2) is smoothed by a size of 1, and the span (3, 7) is smoothed by a size of 2.}
    \label{fig:boundary-smoothing-example}
\end{figure}
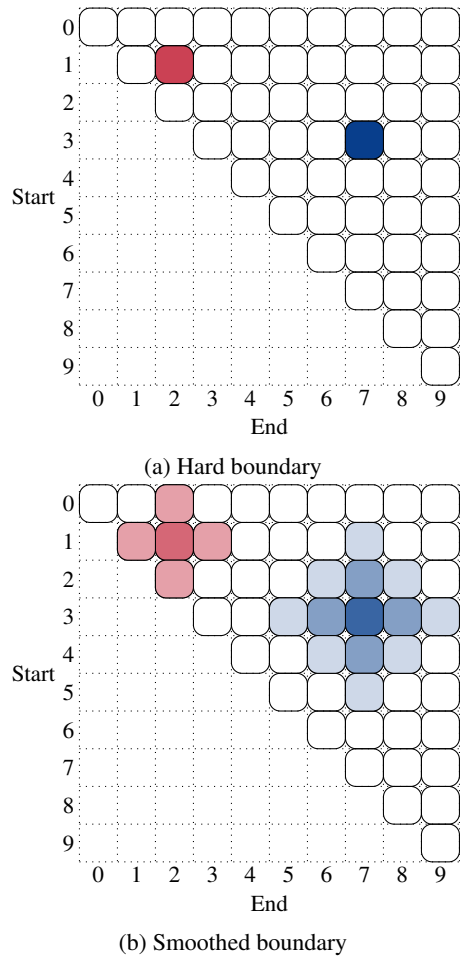

\subsection{Boundary Smoothing} 
Figure~\ref{subfig:hard-boundaries} visualizes the ground truth $y_{ij}$ for an example sentence with two annotated entities. The valid candidate spans cover the upper triangular area of the matrix. In existing NER models, the annotated boundaries are considered to be absolutely reliable. Hence, each annotated span is assigned with the full probability to be an entity, whereas all unannotated spans are assigned with zero probability. We refer to this probability allocation as \emph{hard boundary}, which is, however, probably not the best choice. 

As aforementioned, the entity boundaries may be ambiguous and inconsistent, so the spans surrounding an annotated one deserve a small probability to be an entity. Figure~\ref{subfig:smoothed-boundaries} visualizes $\tilde{y}_{ij}$, the boundary smoothing version of $y_{ij}$. Specifically, given an annotated entity, a portion of probability $\epsilon$ is assigned to its surrounding spans, and the remaining probability $1 - \epsilon$ is assigned to the originally annotated span. With smoothing size $D$, all the spans with Manhattan distance $d$~$(d \leq D)$ to the annotated entity equally share probability $\epsilon / D$. After such entity probability re-allocation, any remaining probability of a span is assigned to be ``non-entity''. We refer to this as \emph{smoothed boundary}.

Thus, the biaffine model can be optimized by the boundary-smoothing regularized cross entropy loss: 
\begin{equation}
    \mathcal{L}_\mathrm{BS} = -\sum_{0 \leq i \leq j < T} \tilde{y}_{ij}^{\mathrm T} \log(\hat{y}_{ij}).
\end{equation}

Empirically, the positive samples (i.e., ground-truth entities) are sparsely distributed over the candidate spans. For example, the CoNLL 2003 dataset has about 35 thousand entities, which represent only 0.93\% in the 3.78 million candidate spans. By explicitly assigning probability to surrounding spans, boundary smoothing prevents the model from concentrating all probability mass on the scarce positive samples. This intuitively helps alleviate over-confidence. 

In addition, hard boundary presents noticeable sharpness between the classification targets of positive spans and surrounding ones, although they share similar contextualized representations. Smoothed boundary provides more continuous targets across spans, which are conceptually more compatible with the inductive bias of neural networks that prefers continuous solutions \citep{hornik1989multilayer}.

\section{Experiments} \label{sec:experiments}
\subsection{Experimental Settings}
\paragraph{Datasets} We use four English NER datasets: CoNLL 2003 \citep{tjong-kim-sang-veenstra-1999-representing}, OntoNotes 5\footnote{\url{https://catalog.ldc.upenn.edu/LDC2013T19}; Data splits follow \citet{pradhan-etal-2013-towards}.}, ACE 2004\footnote{\url{https://catalog.ldc.upenn.edu/LDC2005T09}; Data splits follow \citet{lu-roth-2015-joint}.} and ACE 2005\footnote{\url{https://catalog.ldc.upenn.edu/LDC2006T06}; Data splits follow \citet{lu-roth-2015-joint}.}; and four Chinese NER datasets: OntoNotes 4\footnote{\url{https://catalog.ldc.upenn.edu/LDC2011T03}; Data splits follow \citet{che-etal-2013-named}.}, MSRA \citep{levow-2006-third}, Weibo NER \citep{peng-dredze-2015-named} and Resume NER \citep{zhang-yang-2018-chinese}. Among them, ACE 2004 and ACE 2005 are nested NER tasks, and the others are flat tasks. 

\paragraph{Hyperparameters} For English corpora, we use RoBERTa \citep{liu2019roberta} followed by a BiLSTM layer to produce the contextualized representations. For Chinese, we choose the BERT pretrained with whole word masking \citep{cui2019pretraining}. The BiLSTM has one layer and 200 hidden size with dropout rate of 0.5. The biaffine decoder follows \citet{yu-etal-2020-named}, with the affine layers of hidden size 150 and dropout rate 0.2. We additionally introduce a span width embedding of size 25. Note that the pretrained language models are all of the \texttt{base} size (768 hidden size, 12 layers), and the model is free of any additional auxiliary embeddings; this configuration is relatively simple, compared with those in related work. 

The boundary smoothing parameter $\epsilon$ is selected in $\{0.1, 0.2, 0.3\}$; smoothing size $D$ is selected in $\{1, 2\}$. 

All the models are trained by the AdamW optimizer \citep{loshchilov2018decoupled} with a gradient clipping at L2-norm of 5.0 \citep{pascanu2013difficulty}. The models are trained for 50 epochs with batch size of 48. The learning rate is searched between 1e-3 and 3e-3 on the randomly initialized weights, and between 8e-6 and 3e-5 on the pretrained weights; a scheduler of linear warmup in the first 20\% steps followed by linear decay is applied. 

\paragraph{Evaluation} A predicted entity is considered correct if its type and boundaries exactly match the ground truth. Hyperparameters are tuned according to the $F_1$ scores on the development set, and the evaluation metrics (precision, recall, $F_1$ score) are reported on the testing set. 

\subsection{Main Results}

Table~\ref{tab:english-res} presents the evaluation results on four English datasets, in which CoNLL 2003 and OntoNotes 5 are flat NER corpora, whereas ACE 2004 and ACE 2005 contains a high proportion of nested entities. Compared with previous SOTA systems, our simple baseline (RoBERTa-base + BiLSTM + Biaffine) achieves on-par or slightly inferior performance. Provided the strong baseline, our experiments show that boundary smoothing can effectively and consistently boost the $F_1$ score of entity recognition across different datasets. With the help of boundary smoothing, our model outperforms the best of the previous SOTA systems by a magnitude from 0.2 to 0.5 percentages. 

\begin{table}[!ht]
    \centering \small
    \begin{tabular}{lccc}
        \toprule
        \multicolumn{4}{c}{CoNLL 2003} \\
        \midrule
        Model & Prec. & Rec. & F1 \\
        \midrule
        \citet{lample-etal-2016-neural}     & -- & -- & 90.94 \\
        \citet{chiu-nichols-2016-named}$\dagger$ & 91.39 & 91.85 & 91.62 \\
        \citet{peters-etal-2018-deep}       & -- & -- & 92.22 \\
        \citet{akbik-etal-2018-contextual}$\dagger$ & -- & -- & 93.07 \\
        \citet{devlin-etal-2019-bert}       & -- & -- & 92.8~~ \\
        \citet{strakova-etal-2019-neural}$\dagger$ & -- & -- & 93.38 \\
        \citet{wang-etal-2019-crossweigh}$\dagger$ & -- & -- & 93.43 \\
        \citet{li-etal-2020-unified}        & 92.33 & 94.61 & 93.04 \\
        \citet{yu-etal-2020-named}$\dagger$ & 93.7~~ & 93.3~~ & 93.5~~ \\ 
        Baseline          & 92.93 & 94.03 & 93.48 \\
        Baseline + BS     & 93.61 & 93.68 & \textbf{93.65} \\
        \bottomrule
        \toprule
        \multicolumn{4}{c}{OntoNotes 5} \\
        \midrule
        Model & Prec. & Rec. & F1 \\
        \midrule
        \citet{chiu-nichols-2016-named} & 86.04 & 86.53 & 86.28 \\
        \citet{li-etal-2020-unified}    & 92.98 & 89.95 & 91.11 \\
        \citet{yu-etal-2020-named}      & 91.1~~ & 91.5~~ & 91.3~~ \\
        Baseline          & 90.31 & 92.13 & 91.21 \\
        Baseline + BS     & 91.75 & 91.74 & \textbf{91.74} \\
        \bottomrule
        \toprule
        \multicolumn{4}{c}{ACE 2004} \\
        \midrule
        Model & Prec. & Rec. & F1 \\
        \midrule
        \citet{katiyar-cardie-2018-nested}  & 73.6~~ & 71.8~~ & 72.7~~ \\
        \citet{strakova-etal-2019-neural}$\dagger$ & -- & -- & 84.40 \\
        \citet{li-etal-2020-unified}        & 85.05 & 86.32 & 85.98 \\
        \citet{yu-etal-2020-named}          & 87.3~~ & 86.0~~ & 86.7~~ \\
        \citet{shen-etal-2021-locate}       & 87.44 & 87.38 & 87.41 \\
        Baseline          & 86.67 & 88.42 & 87.54 \\
        Baseline + BS     & 88.43 & 87.53 & \textbf{87.98} \\
        \bottomrule
        \toprule
        \multicolumn{4}{c}{ACE 2005} \\
        \midrule
        Model & Prec. & Rec. & F1 \\
        \midrule
        \citet{katiyar-cardie-2018-nested}  & 70.6~~ & 70.4~~ & 70.5~~ \\
        \citet{strakova-etal-2019-neural}$\dagger$ & -- & -- & 84.33 \\
        \citet{li-etal-2020-unified}        & 87.16 & 86.59 & 86.88 \\
        \citet{yu-etal-2020-named}          & 85.2~~ & 85.6~~ & 85.4~~ \\
        \citet{shen-etal-2021-locate}       & 86.09 & 87.27 & 86.67 \\
        Baseline          & 84.29 & 88.97 & 86.56 \\
        Baseline + BS     & 86.25 & 88.07 & \textbf{87.15} \\
        \bottomrule
    \end{tabular}
    \caption{Results of English named entity recognition. BS means boundary smoothing. $\dagger$ means that the model is trained with both the training and development splits.}
    \label{tab:english-res}
\end{table}

Table~\ref{tab:chinese-res} presents the results on four Chinese datasets, which are all flat NER corpora. Again, boundary smoothing consistently improves model performance against the baseline (BERT-base-wwm + BiLSTM + Biaffine) across all datasets. In addition, our model outperforms previous SOTA by 2.16 and 0.55 percentages on Weibo and Resume NER datasets, and achieves comparable $F_1$ scores on OntoNotes 4 and MSRA. Note that almost all previous systems solve these tasks within a sequence tagging framework; this work adds to the literature by introducing a span-based approach and establishing SOTA results on multiple Chinese NER benchmarks. 

\begin{table}[!ht]
    \centering \small
    \begin{tabular}{lccc}
        \toprule
        \multicolumn{4}{c}{OntoNotes 4} \\
        \midrule
        Model & Prec. & Rec. & F1 \\
        \midrule
        \citet{zhang-yang-2018-chinese} & 76.35 & 71.56 & 73.88 \\
        \citet{ma-etal-2020-simplify}   & 83.41 & 82.21 & 82.81 \\
        \citet{li-etal-2020-flat}       & -- & -- & 81.82 \\
        \citet{li-etal-2020-unified}    & 82.98 & 81.25 & 82.11 \\
        \citet{chen-kong-2021-enhancing} & 79.25 & 80.66 & 79.95 \\
        \citet{wu-etal-2021-mect}       & -- & -- & 82.57 \\
        Baseline          & 82.79 & 81.27 & 82.03 \\
        Baseline + BS     & 81.65 & 84.03 & \textbf{82.83} \\
        \bottomrule
        \toprule
        \multicolumn{4}{c}{MSRA} \\
        \midrule
        Model & Prec. & Rec. & F1 \\
        \midrule
        \citet{zhang-yang-2018-chinese} & 93.57 & 92.79 & 93.18 \\
        \citet{ma-etal-2020-simplify}   & 95.75 & 95.10 & 95.42 \\
        \citet{li-etal-2020-flat}       & -- & -- & 96.09 \\
        \citet{li-etal-2020-unified}    & 96.18 & 95.12 & 95.75 \\
        \citet{wu-etal-2021-mect}       & -- & -- & 96.24 \\
        Baseline          & 95.82 & 95.78 & 95.80 \\
        Baseline + BS     & 96.37 & 96.15 & \textbf{96.26} \\
        \bottomrule
        \toprule
        \multicolumn{4}{c}{Weibo NER} \\
        \midrule
        Model & Prec. & Rec. & F1 \\
        \midrule
        \citet{zhang-yang-2018-chinese} & -- & -- & 58.79 \\
        \citet{ma-etal-2020-simplify}   & -- & -- & 70.50 \\
        \citet{li-etal-2020-flat}       & -- & -- & 68.55 \\
        \citet{shen-etal-2021-locate}   & 70.11 & 68.12 & 69.16 \\
        \citet{chen-kong-2021-enhancing} & -- & -- & 70.14 \\
        \citet{wu-etal-2021-mect}       & -- & -- & 70.43 \\
        Baseline          & 68.65 & 74.40 & 71.41 \\
        Baseline + BS     & 70.16 & 75.36 & \textbf{72.66} \\
        \bottomrule
        \toprule
        \multicolumn{4}{c}{Resume NER} \\
        \midrule
        Model & Prec. & Rec. & F1 \\
        \midrule
        \citet{zhang-yang-2018-chinese} & 94.81 & 94.11 & 94.46 \\
        \citet{ma-etal-2020-simplify}   & 96.08 & 96.13 & 96.11 \\
        \citet{li-etal-2020-flat}       & -- & -- & 95.86 \\
        \citet{wu-etal-2021-mect}       & -- & -- & 95.98 \\
        Baseline          & 95.81 & 96.87 & 96.34 \\
        Baseline + BS     & 96.63 & 96.69 & \textbf{96.66} \\
        \bottomrule
    \end{tabular}
    \caption{Results of Chinese named entity recognition. BS means boundary smoothing.}
    \label{tab:chinese-res}
\end{table}

In five out of the above eight datasets, integrating boundary smoothing significantly increases the precision rate with a slight drop in the recall, resulting in a better overall $F_1$ score. This is consistent with our expectation, because boundary smoothing discourages over-confidence when recognizing entities, which implicitly leads the model to establish a more critical threshold to admit entities. 

Given the use of well pretrained language models, most of the performance gains are relatively marginal. However, boundary smoothing can work effectively and consistently for different languages and datasets. In addition, it is easy to implement and integrate into any span-based neural NER models, with almost no side effects.

\subsection{Ablation Studies}
We perform ablation studies on CoNLL 2003, ACE 2005 and Resume NER datasets (covering flat/nested and English/Chinese datasets), to evaluate the effects of boundary smoothing parameter $\epsilon$ and $D$, as well as other components of our NER system. 

\paragraph{Boundary Smoothing Parameters} We train the model with $\epsilon$ in $\{0.1, 0.2, 0.3\}$ and $D$ in $\{1, 2\}$; the corresponding results are reported in Table~\ref{tab:ablation-parameters}. Most combinations of the two hyperparameters can achieve higher $F_1$ scores than the baseline, which suggests the robustness of boundary smoothing. On the other hand, the best smoothing parameters are different across datasets, which are probably related to the languages/domains of the text, the entity types, and the annotation scheme (e.g., flat or nested NER). Hence, if the best performance is desired for a new NER task in practice, hyperparameter tuning would be necessary. 

\paragraph{Label Smoothing} We replace boundary smoothing with label smoothing in the span classifier. Label smoothing cannot improve, or may even impair the performance of the model, compared with the baseline (see Table~\ref{tab:ablation-parameters}). As aforementioned, we hypothesize that the semantic differences between the typical entity types are quite clear, so it is ineffective to smooth between them. 

\begin{table}[t]
    \centering \small
    \begin{tabular}{lccc}
        \toprule
         & CoNLL & ACE  & Resume \\
         & 2003  & 2005 & NER \\
        \midrule
        Baseline          & 93.48 & 86.56 & 96.34 \\
        \midrule
        BS ($\epsilon$ = 0.1, $D$ = 1) & 93.50 & 86.65 & 96.63 \\
        BS ($\epsilon$ = 0.2, $D$ = 1) & 93.56 & 86.96 & \textbf{96.66} \\
        BS ($\epsilon$ = 0.3, $D$ = 1) & \textbf{93.65} & 86.81 & 96.50 \\
        BS ($\epsilon$ = 0.1, $D$ = 2) & 93.45 & \textbf{87.15} & 96.33 \\
        BS ($\epsilon$ = 0.2, $D$ = 2) & 93.39 & 86.99 & 96.62 \\
        BS ($\epsilon$ = 0.3, $D$ = 2) & 93.57 & 86.71 & 96.28 \\
        \midrule
        LS ($\alpha$ = 0.1) & 93.43 & 86.31 & 96.31 \\
        LS ($\alpha$ = 0.2) & 93.37 & 86.17 & 96.38 \\
        LS ($\alpha$ = 0.3) & 93.26 & 85.65 & 96.26 \\
        \bottomrule
    \end{tabular}
    \caption{Ablation studies of smoothing parameters. $F_1$ scores are reported. BS and LS mean boundary smoothing and label smoothing, respectively.}
    \label{tab:ablation-parameters}
\end{table}

\paragraph{Pretrained Language Models} We test if the performance gain by boundary smoothing is robust to different baselines. For English datasets, we use BERT \citep{devlin-etal-2019-bert} of the \texttt{base} and \texttt{large} sizes, and RoBERTa \citep{liu2019roberta} of the \texttt{large} size (1024 hidden size, 24 layers). It shows that boundary smoothing can consistently increase the $F_1$ scores by 0.1--0.2 and 0.4--0.6 percentages for CoNLL 2003 and ACE 2005, respectively. For Chinese, we use MacBERT \citep{cui-etal-2020-revisiting} of the \texttt{base} and \texttt{large} sizes, and boundary smoothing still performs positively and consistently, with an improvement of 0.2--0.3 percentage $F_1$ scores on Resume NER (see Table~\ref{tab:ablation-structure}). 

It is noteworthy that boundary smoothing achieves performance gains roughly comparable to the gains by switching the pretrained language model from the \texttt{base} size to the \texttt{large} size. This suggests that the effect of boundary smoothing is quite considerable, although the performance improvements seem small in magnitude. 

In addition, our results show that RoBERTa substantially outperforms the original BERT on English NER. This is probably because that (1) RoBERTa is trained on much more data; and (2) RoBERTa focuses on the token-level task (i.e., masked language modeling) by removing the sequence-level objective (i.e., next sentence prediction), hence, it is particularly suitable for within-sequence downstream tasks, e.g., NER. This is also the reason why we choose RoBERTa for our baseline.

\paragraph{BiLSTM Layer} We remove the BiLSTM layer, directly feeding the output of pretrained language model into the biaffine decoder. The results show that this does not change the positive effect of boundary smoothing (see Table~\ref{tab:ablation-structure}). In addition, absence of the BiLSTM layer will result in drops of the $F_1$ scores by about 0.3, 0.5 and 0.1 percentages on the three datasets. 


\begin{table}[t]
    \centering \small
    \begin{tabular}{lccc}
        \toprule
         & CoNLL & ACE  & Resume \\
         & 2003  & 2005 & NER \\
        \midrule
        Baseline                  & 93.48 & 86.56 & 96.34 \\
        \quad + BS                & \textbf{93.65} & \textbf{87.15} & \textbf{96.66} \\
        \midrule
        Baseline w/ BERT-base     & 91.84 & 84.51 & \\
        \quad + BS                & \textbf{92.05} & \textbf{84.95} & \\
        \midrule
        Baseline w/ BERT-large    & 92.92 & 85.83 & \\
        \quad + BS                & \textbf{93.08} & \textbf{86.33} & \\
        \midrule
        Baseline w/ RoBERTa-large & 93.66 & 87.82 & \\
        \quad + BS                & \textbf{93.77} & \textbf{88.02} & \\
        \midrule
        Baseline w/ MacBERT-base  &  &  & 96.41 \\
        \quad + BS                &  &  & \textbf{96.75} \\
        \midrule
        Baseline w/ MacBERT-large &  &  & 96.46 \\
        \quad + BS                &  &  & \textbf{96.75} \\
        \midrule
        Baseline w/o BiLSTM       & 93.13 & 86.22 & 96.24 \\
        \quad + BS                & \textbf{93.30} & \textbf{86.58} & \textbf{96.56} \\
        \bottomrule
    \end{tabular}
    \caption{Ablation studies of model structure. $F_1$ scores are reported. BS means boundary smoothing.}
    \label{tab:ablation-structure}
\end{table}

\section{Further In-Depth Analysis}
\subsection{Over-Confidence and Entity Calibration}
The model performance (evaluated by, e.g., accuracy or $F_1$ score) is certainly important. However, the \emph{confidences} of model predictions are also of interest in many applications. For example, when it requires the predicted entities to be highly reliable (i.e., precision is of more priority than recall), we may filter out the entities with confidences lower than a specific threshold. 

However, \citet{guo2017calibration} have indicated that modern neural networks are poorly calibrated, and typically over-confident with their predictions. By calibration, they mean the extent to which the prediction confidences produced by a model can represent the true correctness probability. We find neural NER models also easy to become miscalibrated and over-confident. We observe that, with the standard cross entropy loss, both the development loss and $F_1$ score increase in the later training stage, which goes against the common perception that the loss and $F_1$ score should change in the opposite directions. This phenomenon is similar to the disconnect between negative likelihood and accuracy in image classification described by \citet{guo2017calibration}. We suppose that the model becomes over-confident with its predictions, including the incorrect ones, which contributes to the increase of loss (see Appendix~\ref{sec:dev-loss} for more details). 

To formally investigate the over-confidence issue, we plot the reliability diagrams and calculate expected calibration error (ECE). In brief, for an NER model, we group all the predicted entities by the associated confidences into ten bins, and then calculate the precision rate for each bin. If the model is well calibrated, the precision rate should be close to the confidence level for each bin (see Appendix~\ref{sec:calibration} for more details). 

\begin{figure}[t]
    \centering
    \begin{subfigure}{\columnwidth}
    \centering
    \includegraphics[width=0.9\columnwidth]{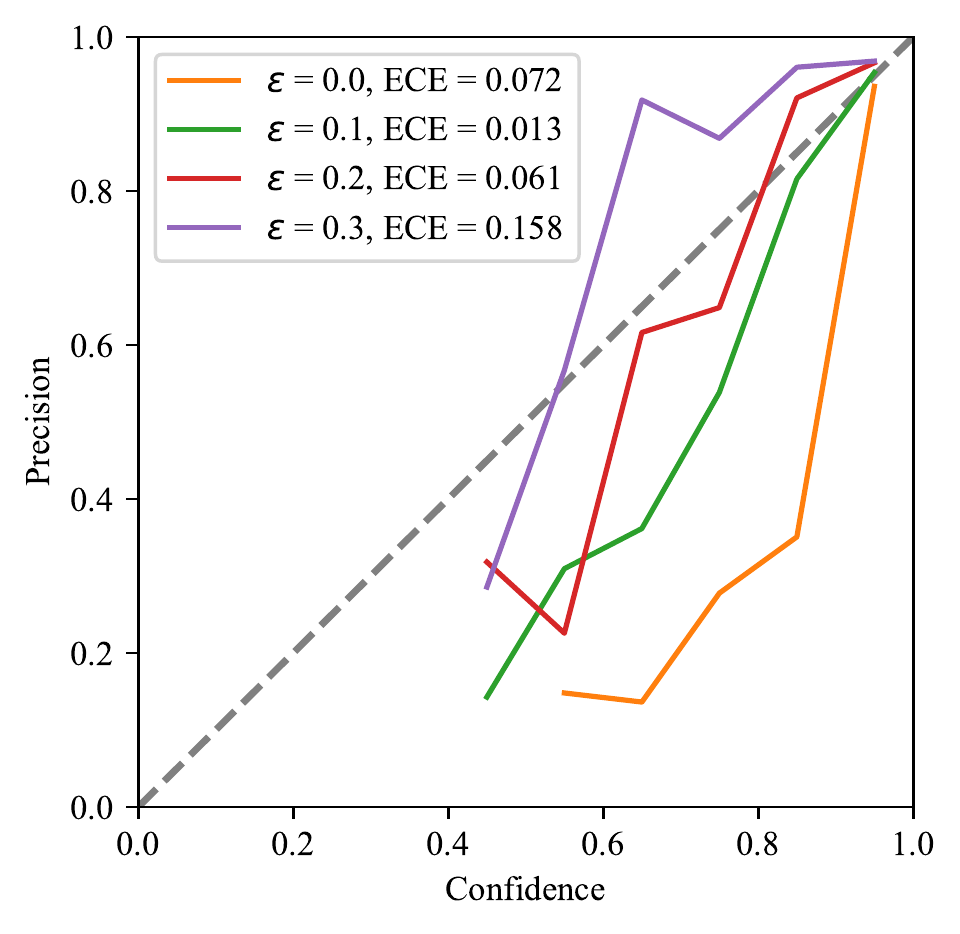}
    \caption{CoNLL 2003}
    \end{subfigure}
    \begin{subfigure}{\columnwidth}
    \centering
    \includegraphics[width=0.9\columnwidth]{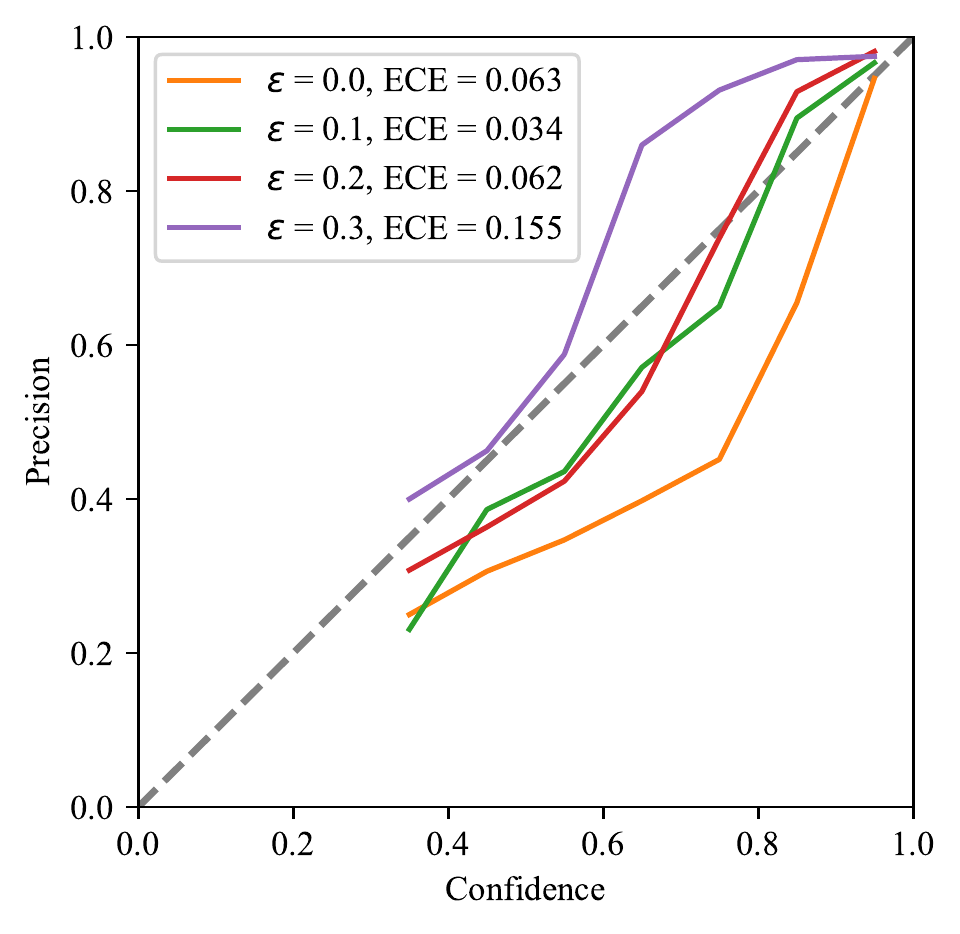}
    \caption{OntoNotes 5}
    \end{subfigure}
    \caption{Reliability diagram of recognized entities on CoNLL 2003 and OntoNotes 5. Results are computed on ten bins.}
    \label{fig:calibration}
\end{figure}

Figure~\ref{fig:calibration} compares the reliability diagrams and ECEs between models with different smoothness $\epsilon$ on CoNLL 2003 and OntoNotes 5. For the baseline model ($\epsilon$ = 0), the precision rates are much lower than corresponding confidence levels, suggesting significant over-confidence. By introducing boundary smoothing and increasing the smoothness $\epsilon$, the over-confidence is gradually mitigated, and shifted to under-confidence ($\epsilon$ = 0.3). In general, the model presents best reliability diagrams when $\epsilon$ is 0.1 or 0.2. In addition, the ECEs of the baseline model are 0.072 and 0.063 on CoNLL 2003 and OntoNotes 5, respectively; with $\epsilon$ of 0.1, the ECEs are reduced to 0.013 and 0.034.

In conclusion, boundary smoothing can prevent the model from becoming over-confident with the predicted entities, and result in better calibration. In addition, as mentioned previously, spans with lower confidences are discarded if they clash with those of higher confidences when decoding. With the better calibration, the model can obtain a very marginal but consistent increase in the $F_1$ score.


\begin{figure*}[t]
    \centering
    \begin{subfigure}{0.3\textwidth}
    \centering
    \includegraphics[width=\textwidth]{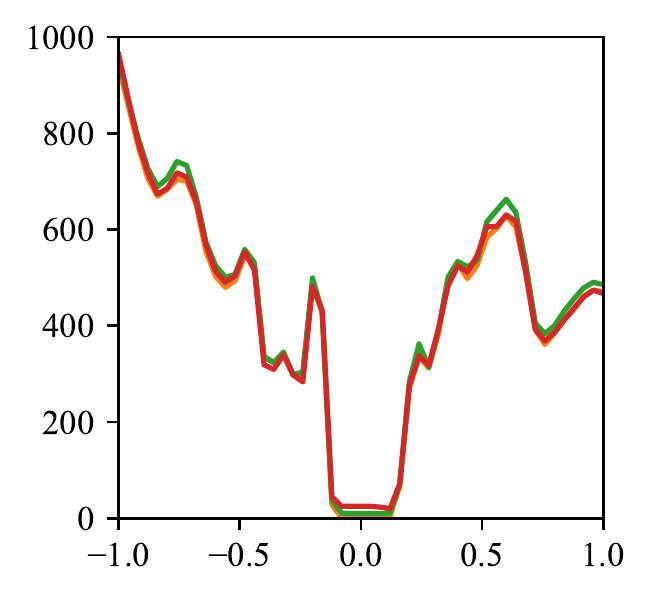}
    \caption{CoNLL 2003, CE}
    \end{subfigure}
    \begin{subfigure}{0.3\textwidth}
    \centering
    \includegraphics[width=\textwidth]{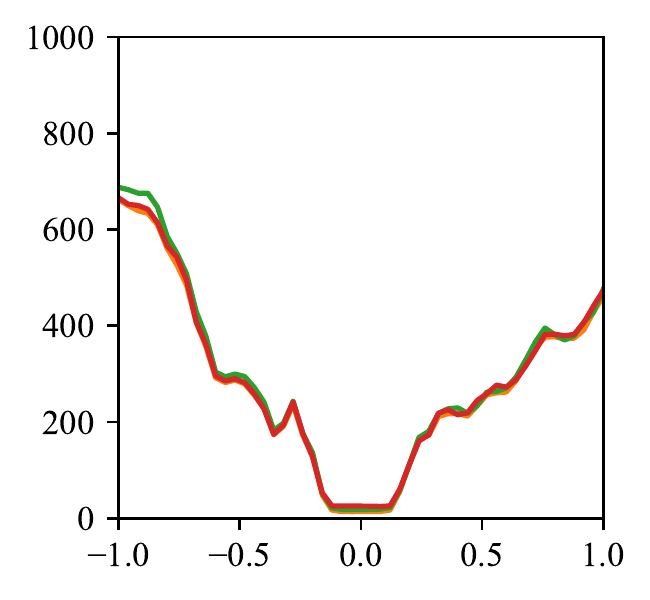}
    \caption{CoNLL 2003, BS ($\epsilon$ = 0.1)}
    \end{subfigure}
    \begin{subfigure}{0.3\textwidth}
    \centering
    \includegraphics[width=\textwidth]{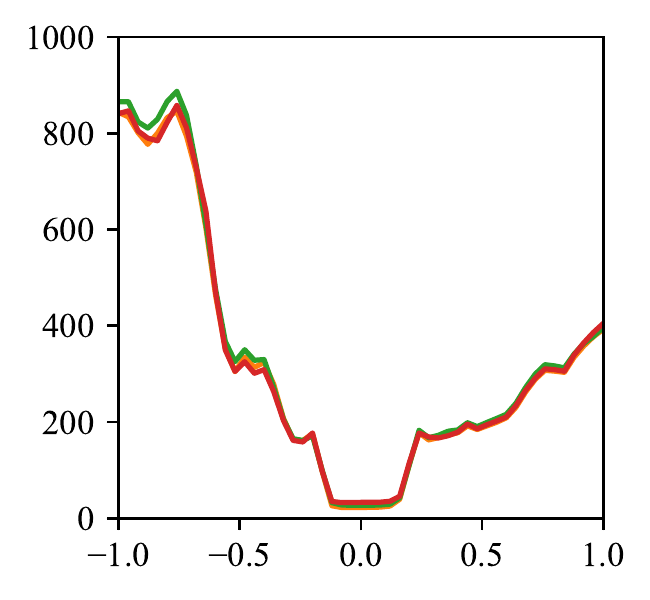}
    \caption{CoNLL 2003, BS ($\epsilon$ = 0.2)}
    \end{subfigure}
    \begin{subfigure}{0.3\textwidth}
    \centering
    \includegraphics[width=\textwidth]{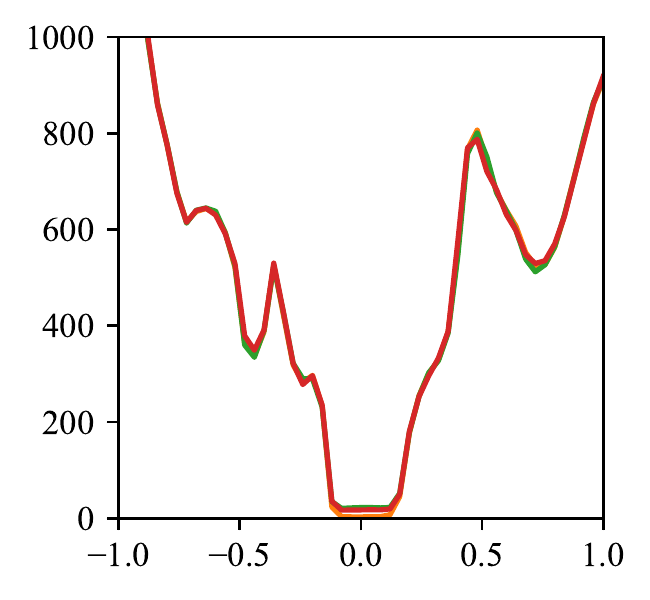}
    \caption{OntoNotes 5, CE}
    \end{subfigure}
    \begin{subfigure}{0.3\textwidth}
    \centering
    \includegraphics[width=\textwidth]{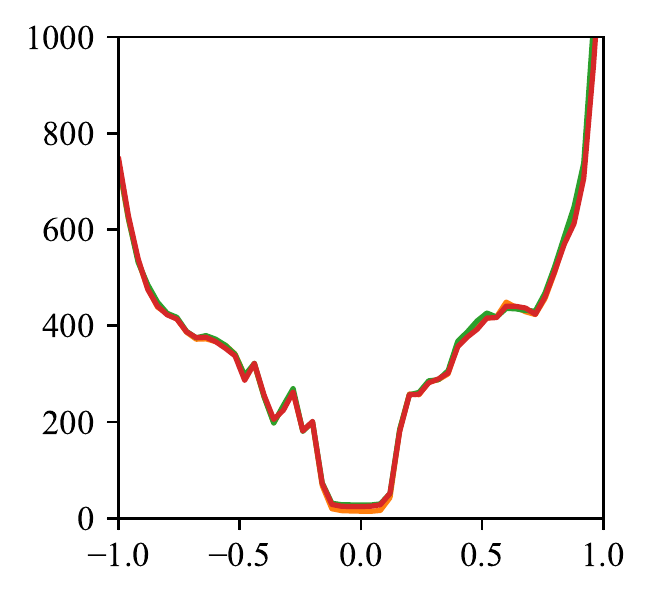}
    \caption{OntoNotes 5, BS ($\epsilon$ = 0.1)}
    \end{subfigure}
    \begin{subfigure}{0.3\textwidth}
    \centering
    \includegraphics[width=\textwidth]{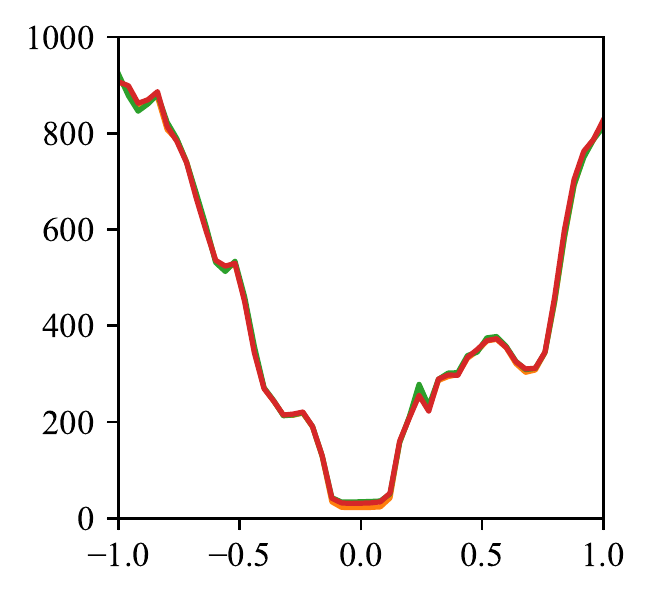}
    \caption{OntoNotes 5, BS ($\epsilon$ = 0.2)}
    \end{subfigure}
    \caption{Visualization of loss landscapes on CoNLL 2003 and OntoNotes 5. Training, development and testing losses are in \textcolor{orange}{orange}, \textcolor{teal}{green} and \textcolor{red}{red}, respectively. CE and BS mean cross entropy and boundary smoothing, respectively.}
    \label{fig:landscape}
\end{figure*}

\subsection{Loss Landscape Visualization}
How does boundary smoothing improve the model performance? We originally conjectured that boundary smoothing can de-noise the inconsistently annotated entity boundaries \citep{lukasik2020does}, but failed to find enough evidence -- the performance improvement did not significantly increase when we injected boundary noises into the training data.\footnote{On the other hand, this cannot rule out the de-noising effect of boundary smoothing, because the synthesized boundary noises are distributed differently from the real noises.} 

As aforementioned, positive samples are very sparse among the candidate spans. Without boundary smoothing, the annotated spans are regarded to be entities with full probability, whereas all other spans are assigned with zero probability. This creates noticeable \emph{sharpness} between the targets of the annotated spans and surrounding ones, although their neural representations are similar. Boundary smoothing re-allocates the entity probabilities across contiguous spans, which mitigates the sharpness and results in more continuous targets. Conceptually, such targets are more compatible with the inductive bias of neural networks that prefers continuous solutions \citep{hornik1989multilayer}. 

\citet{li2018visualizing} have shown that residual connections and well-tuned hyperparameters (e.g., learning rate, batch size) can produce flatter minima and less chaotic loss landscapes, which account for the better generalization and trainability. Their findings provide important insights into the geometric properties of non-convex neural loss functions. 

Figure~\ref{fig:landscape} visualizes the loss landscapes for models with different smoothness $\epsilon$ on CoNLL 2003 and OntoNotes 5, following \citet{li2018visualizing}. In short, for a trained model, a direction of the parameters is randomly sampled, normalized and fixed, and the loss landscape is computed by sampling over this direction (refer to Appendix~\ref{sec:landscape} for more details).

The visualization results qualitatively show that, the solutions found by the standard cross entropy are relatively sharp, whereas boundary smoothing can help arrive at flatter minima. As many theoretical studies regard the flatness as a promising predictor for model generalization \citep{hochreiter1997flat,jiang2019fantastic}, this result may explain why boundary smoothing can improve the model performance. In addition, boundary smoothing is associated with more smoothed landscapes -- the surrounding local minima are small, shallow, and thus easy for the optimizer to escape. Intuitively, such geometric property suggests that the underlying loss functions are easier to train \citep{li2018visualizing}. 

We believe that the sharpness in the span-based NER targets is probably the reason for the sharp and chaotic loss landscape. Boundary smoothing can effectively mitigate the sharpness, and result in loss landscapes of better generalization and trainability. 

\section{Conclusion}
In this study, we propose boundary smoothing as a regularization technique for span-based neural NER models. Boundary smoothing re-assigns entity probabilities from annotated spans to the surrounding ones. It can be easily integrated into any span-based neural NER systems, but consistently bring improved performance. Built on a simple but strong baseline (a \texttt{base}-sized pretrained language model followed by a BiLSTM layer, and the biaffine decoder), our model achieves SOTA results on eight well-known NER benchmarks, covering English and Chinese, flat and nested NER tasks. 

In addition, experimental results show that boundary smoothing leads to less over-confidence, better model calibration, flatter neural minima and more smoothed loss landscapes. These properties plausibly explain the performance improvement. Our findings shed light on the effects of smoothing regularization technique in the NER task. 

As discussed, boundary smoothing typically increases the overall $F_1$ score at the risk of a slight drop in the recall rate; hence, one may be careful to use it for recall-sensitive applications. Future work will apply boundary smoothing to more variants of span-based NER models, and investigate its effect in a broader range of information extraction tasks.

\section*{Acknowledgements}
We thank Yiyang Liu for his efforts in data processing, and the anonymous reviewers for their insightful comments and feedback. This work is supported by the National Natural Science Foundation of China (No. 62106248), Ningbo Science and Technology Service Industry Demonstration Project (No. 2020F041), and Ningbo Public Service Technology Foundation (No. 2021S152).

\bibliography{anthology,custom,references}
\bibliographystyle{acl_natbib}

\newpage
\appendix

\section{Disconnect between Development Loss and $F_1$ Score} \label{sec:dev-loss}
For most machine learning tasks, the desired metric (e.g., accuracy or $F_1$ score) is non-differentiable and thus cannot be optimized via back-propagation. The loss, on the other hand, is a designed differentiable proxy such that minimizing it can increase the original metric. 

\begin{figure}[t]
    \centering
    \begin{subfigure}{\columnwidth}
    \centering
    \includegraphics[width=\columnwidth]{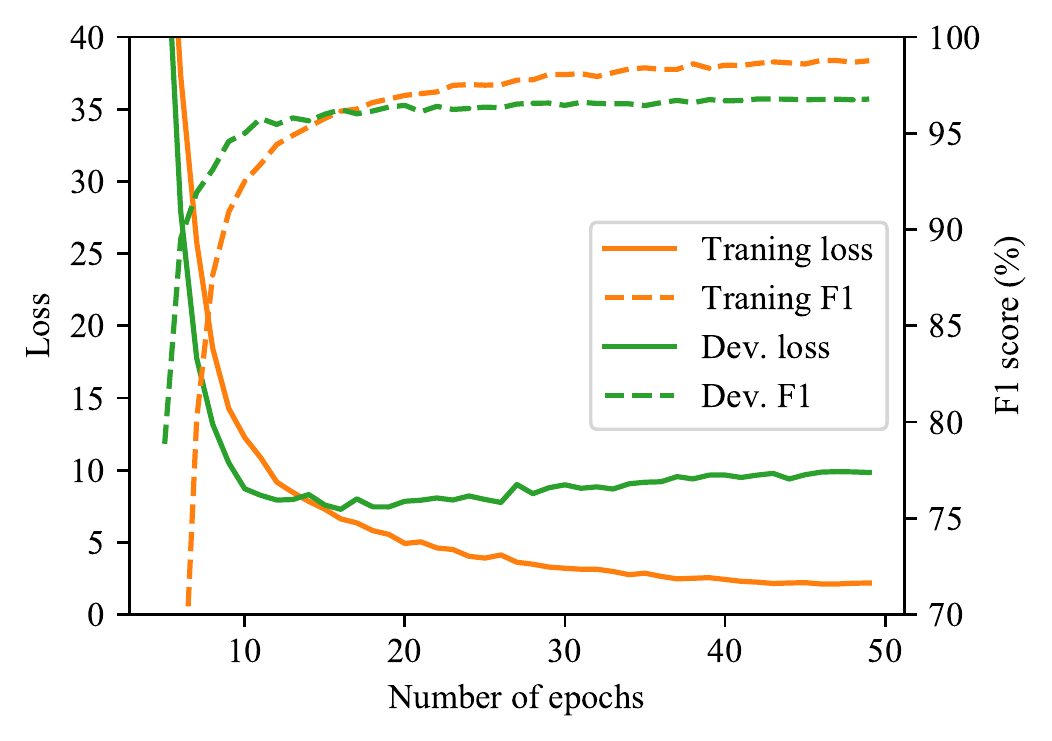}
    \caption{Cross entropy loss}
    \label{subfig:training-curves-ce}
    \end{subfigure}
    \begin{subfigure}{\columnwidth}
    \centering
    \includegraphics[width=\columnwidth]{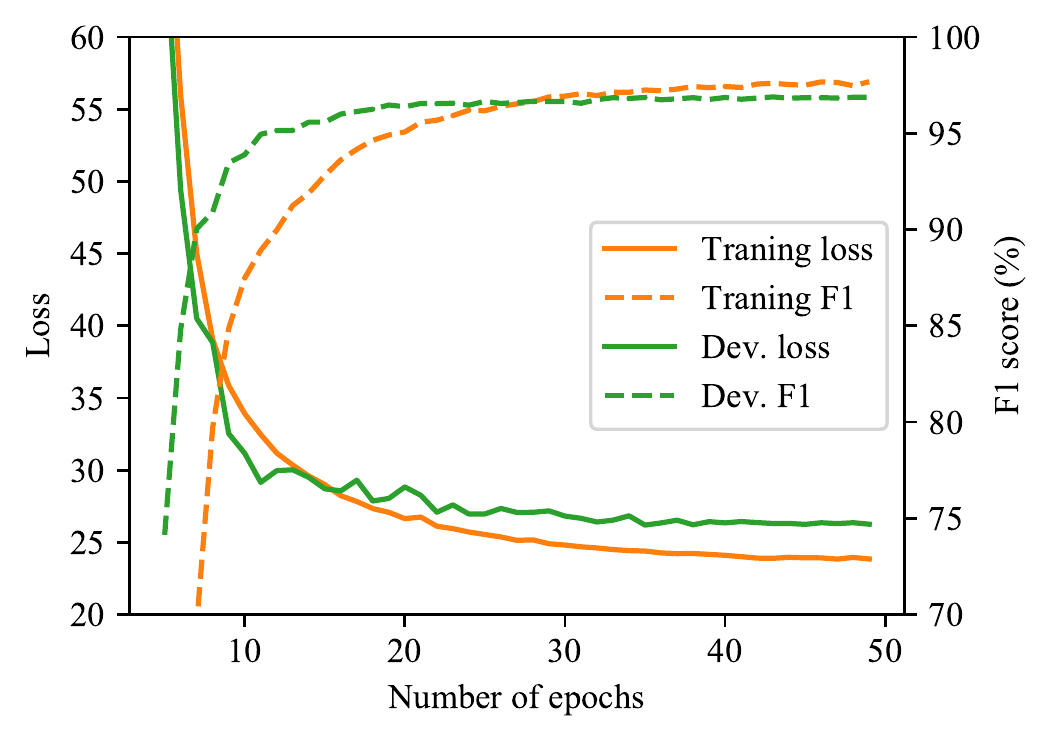}
    \caption{Boundary smoothing loss ($\epsilon$=0.2, $D$=1)}
    \label{subfig:training-curves-bs}
    \end{subfigure}
    \caption{Training/development losses and $F_1$ scores of models with cross entropy loss and boundary smoothing loss on CoNLL 2003. Both the cross entropy loss and corresponding $F_1$ score on the development set experience an ascending trend after about ten epochs, suggesting the existence of over-confidence. However, the boundary smoothing loss on the development set keeps decreasing through the whole training process.}
    \label{fig:training-curves}
\end{figure}

However, as illustrated in Figure~\ref{subfig:training-curves-ce}, when training an NER model by the standard cross entropy loss, although the development $F_1$ score keeps increasing throughout, the development loss also increases in the later stage (e.g., after ten epochs) of the training process. \citet{guo2017calibration} describe this phenomenon as a disconnect -- the neural network overfits to the loss without overfitting to the metric. They regard this as indirect evidence for miscalibration. 

One plausible explanation is that in the later training stage, the model becomes too confident with its predicted outcomes, including both the correct and incorrect ones. Therefore, although slightly more spans are correctly classified on the development set (as the $F_1$ score increases), a small portion of incorrectly classified spans is assigned with much more confidence and contributes to the increase of loss. 

Figure~\ref{subfig:training-curves-bs} presents the curves for boundary smoothing loss. The development loss decreases throughout the training process, opposite to the increasing $F_1$ score. This result suggests that boundary smoothing can help mitigate over-confidence. 

\section{Reliability Diagrams and Expected Calibration Error} \label{sec:calibration}
We generally follow \citet{guo2017calibration}'s approach to plot reliability diagrams and calculate expected calibration error (ECE). 

Given an NER dataset and a model trained on it, denote the gold and predicted entity sets as $\mathcal{E}$ and $\hat{\mathcal{E}}$, respectively; the model produces a confidence $\hat{p}_e$ for each entity $e \in \hat{\mathcal{E}}$. With $K$ confidence interval bins, the predicted entities are grouped such that those with confidences falling into the $k$-th bin constitute a subset:
\begin{equation*}
    \hat{\mathcal{E}}_k = \left\{ e \mid e \in \hat{\mathcal{E}}, \hat{p}_e \in \left( \frac{k-1}{K}, \frac{k}{K} \right] \right\}. 
\end{equation*}

The precision rate (equivalent to the accuracy with regard to a predicted set) of $k$-th group $\hat{\mathcal{E}}_k$ is: 
\begin{equation*}
    \mathrm{Prec}_k = \frac{\vert \hat{\mathcal{E}}_k \cap \mathcal{E} \vert}{\vert \hat{\mathcal{E}}_k \vert},
\end{equation*}

and the corresponding average confidence is: 
\begin{equation*}
    \mathrm{Conf}_k = \frac{\sum_{e \in \hat{\mathcal{E}}_k} \hat{p}_e}{\vert \hat{\mathcal{E}}_k \vert}. 
\end{equation*}

The reliability diagrams plot $\mathrm{Prec}_k$ against $\mathrm{Conf}_k$ for $k = 1,2,\dots,K$. ECE is estimated by the weighted average of absolute difference between $\mathrm{Prec}_k$ and $\mathrm{Conf}_k$: 
\begin{equation*}
    \mathrm{ECE} = \sum_{k=1}^{K} \frac{\vert \hat{\mathcal{E}}_k \vert}{\vert \hat{\mathcal{E}} \vert} \cdot \bigg\vert \mathrm{Prec}_k - \mathrm{Conf}_k \bigg\vert
\end{equation*}

By definition, a perfectly calibrated model will have $\mathrm{Prec}_k = \mathrm{Conf}_k$ for $k = 1,2,\dots,K$. In this case, the reliability diagrams should lie along the identity line, and ECE equals to 0. 

\section{Loss Landscape Visualization} \label{sec:landscape}
We generally follow \citet{li2018visualizing}'s approach to visualize the loss landscape. 

Given a trained model of parameters $\theta^\star$, we sample a random direction $\delta$ from a normal distribution, and rescale it by: 
\begin{equation*}
    \delta_i \leftarrow \frac{\Vert \theta^\star_i \Vert}{\Vert \delta_i \Vert} \delta_i, 
\end{equation*}

where $\delta_i$ is the $i$-th weight of $\delta$.\footnote{\citet{li2018visualizing} use filter-wise normalization for convolutional networks, whereas our models have no convolutional layers, so we simplify it as weight-wise normalization.} On a data set/split $\mathcal{D}$, the loss landscape plots the function: 
\begin{equation*}
    f(\alpha) = L(\mathcal{D}; \theta^\star + \alpha \delta),
\end{equation*}

where $L(\mathcal{D}; \theta)$ is the average loss value (in the evaluation mode) on $\mathcal{D}$ if the model takes parameters of $\theta$. In practice, we evenly sample 51 points in the interval $[-1, 1]$ for $\alpha$, and plot the loss values against $\alpha$. 

\end{document}